\documentclass[10pt,twocolumn,letterpaper]{article}
\pdfoutput=1
\usepackage{cvpr}
\usepackage{times}
\usepackage{epsfig}
\usepackage{graphicx}
\usepackage{amsmath}
\usepackage{amssymb}
\usepackage{algorithm,algorithmic}
\usepackage[toc,page,title,titletoc]{appendix}

\usepackage[breaklinks=true,bookmarks=false]{hyperref}

\cvprfinalcopy 

\def\cvprPaperID{****} 

\begin{document}

\title{$l_1$-regularized Outlier Isolation and Regression}

\author{ Sheng Han\\
Department of Electrical and Electronic Engineering, The University of Hong Kong, HKU\\
Hong Kong, China\\
{\tt\small sheng4151@gmail.com}
\and
Suzhen Wang \\
Department of Information Engineering, The Chinese University of Hong Kong, CUHK\\
Hong Kong, China\\
{\tt\small ws012@ie.cuhk.edu.hk}
\and
Xinyu Wu \\
Shenzhen Institutes of Advanced Technology, CAS \\
ShenZhen, China\\
{\tt\small xy.wu@siat.ac.cn}
}

\maketitle

\begin{abstract}
   This paper proposed a new regression method called $l_1$-regularized outlier isolation and regression (LOIRE) as well as an efficient algorithm to solve it based on the idea of block coordinate descent. Additionally, assuming outliers are gross errors following a Bernoulli process, this paper also proposed a Bernoulli estimate which, in theory, should be very accurate and robust by completely removing every possible outliers. However, the Bernoulli estimate is not easy to achieve but with the help of the proposed LOIRE, it can be approximately solved with a guaranteed accuracy and an extremely high efficiency compared to current most popular robust-estimation algorithms. Moreover, LOIRE can be further extended to realize robust rank factorization which is powerful in recovering low-rank component from high corruptions. Extensive experimental results showed that the proposed method outperforms state-of-the-art methods like RPCA and GoDec in the aspect of computation speed with a competitive performance.
\end{abstract}

\section{Introduction}

Most robust estimates like least absolute deviate \cite{powell1984least} realized by ADMM \cite{boyd2011distributed} and MM-estimator \cite{huber2011robust} are able to give us accurate estimates but they are still computationally intensive which in a certain level prevents these robust estimates from wide use.
In this paper, we aimed to propose an efficient robust estimate called $l_1$-regularized outlier isolation and regression, LOIRE in short, with a very parsimony formulation.
\begin{equation}
\begin{split}
\label{L1Model}
<\hat{x}, \hat{b}> = \min_{x,b} \|b\|_1
\quad s.t.~ \|y-Ax-b\|_2\le t,
\end{split}
\end{equation}
where $A$ denotes the measurement system, $y$ denotes the measured data, $x$ denotes the unknown signal to be determined, $b$ denotes the outlier vector and $t$ to be a nonnegative value.

In fact, LOIRE can be derived by assuming the noise is a mixture of Gaussian noise and Laplace noise. However, considered outliers are large errors with a totally random occurrence, it is more appropriate to assume their occurrences follow a Bernoulli process. With the inevitable small operation noises in the measurement system, we can eventually derive a Bernoulli estimate as follows:
\begin{equation}
\begin{split}
\label{L0Model}
<x^0,b^0> = \min_{x,e} \|b\|_0
\quad s.t.~ \|y-Ax-b\|_2\le t,
\end{split}
\end{equation}
where $<x^0, b^0>$ denotes the optimal solution pair under the Bernoulli assumption.

Obviously, the Bernoulli estimate is difficult to solve and the LOIRE is less accurate. Fortunately, this paper found a way with a theoretical explanation in a certain level to achieve an estimate which combines the Bernoulli estimate's accuracy with the LOIRE's efficiency.

Due to the parsimony and effectiveness of the LOIRE method, this paper further extended this method to realize robust rank factorization which can be applied to recover low-rank structures from massive contaminations, such as background modeling, face recognition and so on.
Assuming that a data matrix $Y\in R^{m\times n}$ consists of two parts: the low-rank part and the contamination part. Then to correctly recover the low rank component equals to detect all the contaminations that can be treated as outliers. Below is the proposed robust factorization model based on LOIRE:
\begin{equation}
\label{RobustRankFactorization}
\begin{split}
\{\hat{A},\hat{X}\}
&=\mathop{\arg\min}_{A,X} \frac{\lambda}{2} \|Y-AX-B\|^2_2 + \|B\|_1.\\
s.t.~&\quad\|A_{\cdot,j}\|_2=1,~ ~\forall j=1,...,n,
\end{split}
\end{equation}
where matrix $A\in R^{m\times r}$ can be understood as a dictionary which contains all the information about the low rank structure, $A_{\cdot,j}$ indicates the $j$-th column of matrix $A\in R^{m\times r}$. Each column of $X \in R^{r\times n}$ denotes a coefficient vector for each column of $Y$. The products $AX$ represents the low-rank component of $Y$ while $B$ reflects the contamination component. Experiments both on simulation data and real image data would verify the high efficiency and strong robustness of this method compared to state-of-the-art approaches like RPCA and GoDec.

\section{Related Work}
In robust statistics, least absolute deviations (LAD) regression \cite{powell1984least} was proposed decades ago and has been used extensively in statistics but it lacks efficiency especially when it comes to deal with large datasets. Recently, Boyd et al. \cite{boyd2011distributed} applied the ADMM algorithm to solve LAD fitting which greatly accelerate the computation speed, but it lacks stability and fluctuate its efficiency according to different dataset. Other popular robust regression methods with fast implementations like fast-LTS \cite{Fast-LTS2006}, fast S-estimator \cite{salibian2006fast}, MM-estimator \cite{huber2011robust} and to name a few, are still facing the same problem of less efficiency, thus making them less practical to complicate real-life problems in a certain level.


To the best of the authors' knowledge, the most effective low-rank recovery methods are \cite{RPCA,wright2009robust,zhou2011godec}, all of them combine strong robustness with high efficiency. For RPCA method \cite{wright2009robust}, its fastest algorithm has been presented later in paper \cite{lin2010augmented}, which is called inexact ALM (IALM). For GoDec \cite{zhou2011godec}, it also published its fastened algorithm in \cite{SSGoDec}, which is called SSGoDec. We would show it later that the proposed rank factorization method based on LOIRE outperforms these state-of-the-art methods in terms of efficiency with a competitive robustness.

\section{$l_1$-regularized Outlier Isolation and Regression}
In this paper, we consider a measurement system $A$ which is has a probability $p$ with $p<\frac{1}{2}$ to be attacked by gross errors. To be practical, we also need to consider the dense and normal operation noises besides the gross errors.
So the measurement process can be expressed in a mathematical form:
\begin{equation}
\begin{split}
y = Ax + b + e\\
\end{split}
\end{equation}
where $y$ it the observation through $A$, $b$ denotes the outlier vector and $e$ denotes a dense Gaussian noise.


By adding the penalty term $\|b\|_1$ to the least mean squares on $e$, we can eventually derive an estimation model for $x$ as follows:
\begin{equation}
\begin{split}
&\min_{e, b} \|e\|_2 + \mu \|b\|_1\\
&s.t.~  \mu > 0, y =Ax + e + b,
\end{split}
\end{equation}
which in fact has an equivalent form as problem (\ref{L1Model}).

\subsection{Alternative Direction Descent Algorithm for LOIRE}
In this subsection, we turn our attention to how to the LOIRE problem (\ref{L1Model}) which is convex but non-derivative. In truth, this problem can be re-formulated as:
\begin{equation}
\min_{x,b}  \frac{\lambda}{2} \|y-Ax-b\|_2^2 + \|b\|_1
\end{equation}
with $\lambda >0$.
Based on the idea of block coordinate descent, we can then derive an efficient algorithm for this problem.
Firstly, fix $b$ and optimize $x$ only, we can get the first convex subproblem:
\begin{equation}
\label{Sub1}
\hat{x} = \arg\min_{x} \|y-Ax-b\|^2_2;
\end{equation}
and given $A$ has full column rank, otherwise, we can apply Moore-Penrose pseudoinverse and then the solution for this sub-problem is:
\begin{equation}
x = (A^TA)^{-1}A^T(y-b).
\end{equation}
Fixed $x$ and optimize $b$ only, we can get the second convex subproblem:
\begin{equation}
\label{Sub2}
\hat{b} = \arg\min_{b}  \|b\|_1+\frac{\lambda}{2} \|y-Ax-b\|^2_2.
\end{equation}
The second subproblem is essentially a lasso problem and paper \cite{Lasso} implies the solution as follows:
\begin{equation}
<\hat{b}_i> = sign((y-Ax)_i)\left(|(y-Ax)_i|-\frac{1}{\lambda}\right)_+.
\end{equation}

Based on these above, we thus can give the alternative direction descent algorithm (ADDA) as shown in Algorithm \ref{alg1}.

\begin{algorithm}[h]
\caption{\label{alg1} Alternative Direction Descent Algorithm for LOIRE}
\begin{algorithmic}[1]
\REQUIRE The vector $y$ and matrix $A$
\ENSURE
\STATE Initialization:\quad $b_0=\vec{0}$,\quad $k =0$;
\WHILE {Not convergent}
\STATE $x_{k+1} = (A^TA)^{\dag}A^T(y-b_k)$.
\STATE let $y_{k+1} = y-Ax_{k+1}$.
\STATE $b_{k+1}= sign[y_{k+1}]\circ\left[|y_{k+1}|-\frac{1}{\lambda}\vec{1}\right]_+$
\STATE k = k+1
\ENDWHILE
\RETURN $x,b$
\end{algorithmic}
\end{algorithm}
where " $\circ$ " is Hadamard product, i.e. entrywise production. The algorithm stops when $b_{k+1}-b_{k}<\epsilon$ for a given small and positive $\epsilon$.

\subsection{Convergence of ADDA}
In this subsection, we aim to show that the above algorithm will converge to an optimal solution.
First we showed the following sequence
\begin{equation}
\{(x_0,b_0),(x_1,b_0),(x_1,b_1),..., (x_k,b_{k-1}),(x_k,b_k),...\}
\end{equation}
converges to a fixed point, then we would show that this fixed point is actually an optimal solution for the LOIRE problem.

Let
\begin{equation}
f(x,b)=\|b\|_1 +\frac{\lambda}{2}|y-Ax-b\|_2^2,
\end{equation}
according to AVOM, we would have
\begin{equation}
f(x_k,b_k)\ge f(x_{k+1},b_k)\ge f(x_{k+1},b_{k+1}) \ge...> -\infty
\end{equation}, so this sequence must converge to a certain fixed point denoted as $(x_0, b_0)$.

Now we came to the second part of the proof: to show $(x_0,b_0)$ is actually an optimal solution.
From above, we could have:
\begin{equation}
f(x_0 +\Delta x, b_0) \le f(x_0 , b_0),~ f(x_0 , b_0+\Delta b ) \le f(x_0 , b_0).
\end{equation}
Then for any convex combination of $(x_0 +\Delta x, b_0)$ and $(x_0 , b_0+\Delta b)$, we should have:
\begin{equation}
\begin{split}
&f(x_0 +\lambda \Delta x, b_0 +(1-\lambda)\Delta b)\le \lambda f(x_0 +\Delta x, b_0)\\ &+(1-\lambda)f(x_0 , b_0+\Delta b )\le f(x_0 , b_0).
\end{split}
\end{equation}
So $(x_0,b_0)$ is actually a local optimal point but since $f(x,b)$ is a convex function, the local optimal point should also be a global one. Thus we finished our proof.

\section{Bernoulli Estimation Model}
In this section, we aimed to arrive at a Bernoulli estimation model (BEM) which is based on Bernoulli distribution assumption for outliers in the observations.

\subsection{Notations}
Let $\mathcal{I}$ be an index set, then $I^c$ and $|\mathcal{I}|$ denote its complementary set and its cardinality respectively.
Let $X$ be a matrix (or a vector), then $X_{\mathcal{I}}$ denotes a submatrix (vector) formed from the rows of $X$ indexed by the elements in $\mathcal{I}$. If $\mathcal{I} = \{i\}$, the notation can be simplified as $X_i$ indicating the $i$-th row of $X$.

\subsection{Bernoulli Estimation Model}
Let $\mathcal{B}(p)$ denote a Bernoulli distribution with an outlier appears with probability $p$. Let $\textbf{1}$ denotes an outlier indicator of $y$: $\textbf{1}_i =0$ indicates that $y_i$ is a normal measurement, otherwise it is an outlier.
If $y_i$ is an outlier with probability $p$ then we can have $\textbf{1}_i\sim \mathcal{B}(p)$:
\begin{equation}
\label{MixtureNoise}
\begin{split}
\textbf{1}_i~:=
\begin{cases}
0, &\quad p  >\frac{1}{2}\\
1,&\quad 1-p,
\end{cases}
\end{split}
\end{equation}
where $p>\frac{1}{2}$ is a necessary guarantee to make a successful estimate.
Let $\mathcal{I}$ be an index set of normal entries of $y$, that is $\forall i\in I, \textbf{1}_{i}=0$, then $\mathcal{I}^c$ denotes the index set of outliers. In other words, we cay say $\mathcal{I}$ is an index set of normal measurements if and only if given a vector $e\in R^m$ satisfying
\begin{equation}
\label{condition1}
\|e\|_2 \le t,
\end{equation}
for some positive real number $t$, then
\begin{equation}
\begin{split}
\label{indicator}
&y_{\mathcal{I}} = A_{\mathcal{I}}x + e_{\mathcal{I}}\\
&y_i\neq A_ix +e_i,\forall i \in \mathcal{I}^c.
\end{split}
\end{equation}
Obviously $e\in R^m$ reflects the acceptable noise in measurements.

According to formula (\ref{indicator}), we can have
\begin{equation}
|I^c| = \|y-Ax-e\|_0,
\end{equation}
with $e$ satisfying condition (\ref{condition1}).
%
Given a specific but unknown $x$, then according to the Bernoulli distribution, we can determine the probability for $y$ with $|I^c|$ outlier entries:
\begin{equation}
P(y|x) = p^{m-\|y-Ax-e\|_0}(1-p)^{\|y-Ax-e\|_0}.
\end{equation}
Then apply the usual maximum log-likelihood method to the above formula, given $p >\frac{1}{2}$, the Bernoulli estimation model is derived as follows:
\begin{equation}
\begin{split}
&<x^0,e^0>=\min_{x, e} \|y-Ax-e\|_0\\
&s.t. \quad\|e\|_2 \le t,
\end{split}
\end{equation}
which has an equivalent formula as follows:
\begin{equation}
\label{L0Model}
\begin{split}
&<x^0,b^0> = \arg\min_{x,b} \|b\|_0 \\
&s.t. \quad\|y-Ax-b\|_2 \le t.
\end{split}
\end{equation}
Thus we obtained the BEM regression model.

\subsection{Relation between BEM and LMS}
\noindent \textbf{Proposition 1} Assume $(x^0,b^0)$ is the optimal solution of problem (\ref{L0Model}) and $\mathcal{I}$ is the support set of $b^0$, then this optimal solution can be equally obtained by solving the following problem of least mean squares instead, i.e. $Ax^0 = Ax^{\prime}$, where
\begin{equation}
x^{\prime} = \arg\min_{x}\|y_{\mathcal{I}^c} - A_{\mathcal{T}^c}x\|_2^2
\end{equation}
The insight that this Proposition conveys to us is: if we can localize all the outliers in advance, then we can apply least mean squares estimation method to these uncorrupted measurements in order to get a Bernoulli estimation for $x$.
The detailed mathematical proof is shown in the Appendices.

\subsection{Approximate Bernoulli Estimate}
For LOIRE regression, it is efficient but it still suffers a slight deviation caused by outliers; for Bernoulli estimate, it is accurate but hard to compute.
Fortunately, inspired by Proposition 1, there is simple way to combine the accuracy of Bernoulli estimate with the efficiency of the LOIRE:
firstly use LOIRE to detect the localization of the outliers;
then remove the entries corrupted by outliers in $y$;
Lastly apply the least mean squares on the cleaned observation.

From Proposition 1, we can see, if LOIRE succeeds in detecting all the outliers, then the above steps would give a Bernoulli estimate. As for algorithm efficiency, the above process only append the "least mean squares" step that is known to be fast to LOIRE, therefore, the above process is efficient and improves accuracy in a certain level.

\section{Rank Factorization based on LOIRE}
\textbf{Notations}: for a matrix $Y$, let $Y_{\cdot,i}$ denote the $i$-th column of matrix $Y$.

Generally, given a matrix $Y\in R^{m\times n}$ with rank less than or equal to $r$, then it can be represented as a product of two matrices, $Y = AX$ with $A\in R^{m\times r}$ and $X\in R^{r\times n}$.
However, in this section, we considered to recover a low-rank component of a seriously contaminated matrix using a robust rank factorization method based on LOIRE.

Before we start the derivation of a new robust rank factorization model based on LOIRE, we should make the following two things clear: one is each column of a contaminated matrix $Y$ should have equal chance to be corrupted; the other is the low-rank component of $Y$ still can be appropriately represented by $AX$. Below is the detailed derivation:

First of all, assuming matrix $A$ is known, we applied LOIRE to each column of $Y$:
\begin{equation}
\label{FrameOpti}
\begin{split}
\min_{B_{\cdot, i},X_{\cdot, i}}\|B_{\cdot, i}\|_1 + \frac{\lambda}{2}\|Y_{\cdot, i} - AX_{\cdot, i}-B_{\cdot, i}\|^2_2, \forall i = 1,2,...,n,
\end{split}
\end{equation}
with $B_{\cdot, i}$ denotes an outlier vector for $i$-th column.
To be concise, we can re-represent formula (\ref{FrameOpti}) in the following form:
\begin{equation}
\min_{B,X} \|B\|_1 +\frac{\lambda}{2}\|Y-AX-B\|_2^2.
\end{equation}

In fact, matrix $A$ is generally unknown, then a simple way to find a most appropriate matrix $A$ that fits the problem is to search one that minimizes the above optimization problem. Thus the optimization problem becomes:
\begin{equation}
\begin{split}
\label{BackgroundModeling}
\mathop{\min}_{A}\mathop{\min}_{B,X}\|B\|_1 +\frac{\lambda}{2} \|Y - AX-B\|^2_2, \\
\end{split}
\end{equation}
To ensure a unique solution for matrix $A$ and $X$, we would like to add a regularization constraint to matrix $A$, that is for each column of $A$, it should have a unit length:
\begin{equation}
\|A_{\cdot,j}\|_2=1, \forall i =1,...,r.
\end{equation}
Eventually, we got the proposed robust rank factorization model.

\subsection{Algorithm for Robust Rank Factorization}
After we derived the robust rank factorization model, we should come to focus on its solution algorithm. Similarly as ADDA, we can split the original problem into two subproblems:
Fix matrix $B$, we can get the first subproblems:
\begin{equation}
\label{form6}
\begin{split}
\{A,X\}
&=\mathop{\arg\min}_{A,X} \|Y - AX-B\|^2_2.\\
s.t.&\quad\|A_{\cdot,j}\|_2=1.
\end{split}
\end{equation}
Fix matrix $A$ and $X$, we can get the second subproblem:
\begin{equation}
\label{form6}
\begin{split}
\{B\}
&=\mathop{\arg\min}_{B} \|B\|_1 + \frac{\lambda}{2}\|Y - AX-B\|^2_2.\\
\end{split}
\end{equation}
The solution for the first subproblem is:
\begin{equation}
\begin{split}
A=U[1:r], \quad X = (\Sigma V^T)(1:r),
\end{split}
\end{equation}
where $(Y-B)= U\Sigma V^T$, i.e. the singular value decomposition with $U[1:r]$ implies to take the first $r$ columns of matrix $U$ and $(\Sigma V^T)(1:r)$ takes the first $r$ rows of matrix $(\Sigma V^T)$.
The solution for the second subproblem is:
\begin{small}
\begin{equation}
\begin{split}
B_{ij} = sign((Y-AX)_{ij})\left(|(Y-AX)_{ij}|-\frac{1}{\lambda}\right)_+,
\end{split}
\end{equation}
\end{small}
where $B_{ij}$ denotes an entry in the $i$-th row and $j$-th column of matrix $B$.

Also similar to ADDA, we can derive an alternative matrix descent algorithm (AMDA) in Algorithm \ref{backModeling} to solve the robust rank factorization in algorithm. Following the same line of the proof for ADDA, one can proof that AMDA will converge to a global optimal solution.

\begin{algorithm}[h]
\caption{\label{backModeling} Alternative Matrix Descent Algorithm for Robust Rank Factorization}
\begin{algorithmic}[1]
\REQUIRE Matrix $Y$
\ENSURE The matrix $A$, $X$,$B$
\STATE Initialization:$k = 0, B_0=\textbf{O}$;
\WHILE {Not converged}
\STATE let $(Y-B_k) = U_k\Sigma_kV_k^T$.
\STATE $A_k = U_k[1:r]$, $X_k = (\Sigma_k V_k^T)(1:r)$.
\STATE let $Y_k = Y-A_kX_k$.
\STATE $B_{k+1} = sign[Y_k]\circ\left[|Y_k|-\frac{1}{\lambda}11^T\right]_+$
\STATE $k = k + 1 $;
\ENDWHILE
\RETURN $B$,$AX$
\end{algorithmic}
\end{algorithm}
The AMDA algorithm stops if $\|B_{k+1} -B_{k}\|_F \le \epsilon$ given a small convergence tolerance $\epsilon >0$.

\section{Experiments}
The experiments is run by Matlab on a laptop with Intel i7 CPU (1.8G) and 8G RAM.
All the reported results are direct outcomes of their corresponding algorithms without any post processing.
\subsection{Approximate Bernoulli Estimate based on LOIRE}

In this part, we will compare the proposed robust regression method with the most famous models in the area of robust regression. Algorithms chosen for each regression model to be compared to are all speeded up versions proposed in recent years. For LAD, we use ADMM proposed by Boyd et al. \cite{boyd2011distributed}; for S-estimation, we use the latest fastened algorithm proposed in paper \cite{salibian2006fast}; for least-trimmed-squares regression we use the latest fastened algorithm proposed in paper \cite{Fast-LTS2006}; MM-estimation applied here is also a fastened algorithm proposed in \cite{huber2011robust}.
We conduct the comparison experiments on data sets from \cite{ROUSSEEUWdatasets}.

\begin{figure}[t]
\begin{center}
   \includegraphics[width=8cm,height=12cm]{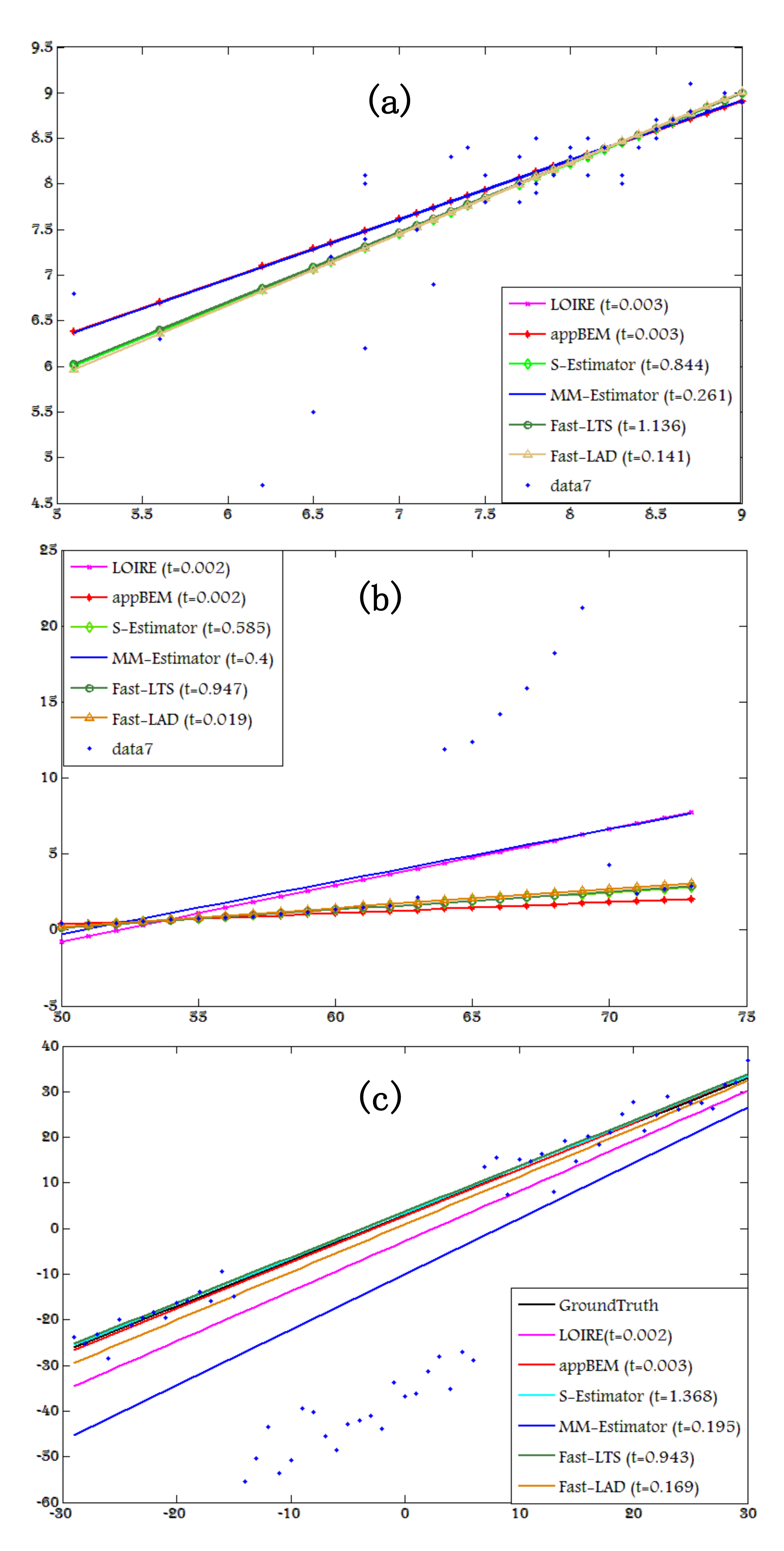}
\end{center}
\vspace{-1em}
   \caption{\label{fig:comparison}\textbf{Comparison Experiment}: Data in the top two graphs is cited from a public dataset \cite{ROUSSEEUWdatasets}; Data in the bottom graph is created by authors with a dark line showing the ground truth. In (a), Lines of LOIRE, appBEM and MM-estimate overlap with each achieves a very accurate estimate. The proposed LOIRE and appBEM achieve the highest efficiency. In (b), Lines of LOIRE and MM-estimate group together and the rest lines form a bunch. The proposed appBEM achieves the highest efficiency with accuracy. In (c) appBEM overlaps with the ground truth with a high efficiency.}
\end{figure}

Fig.\ref{fig:comparison} shows that the approximate Bernoulli estimate holds the highest efficiency with a competitive robustness.
%

\subsection{Robust Rank Factorization}

\subsubsection{Simulations}

\begin{figure*}
\begin{center}
\includegraphics[width=16cm,height=5cm]{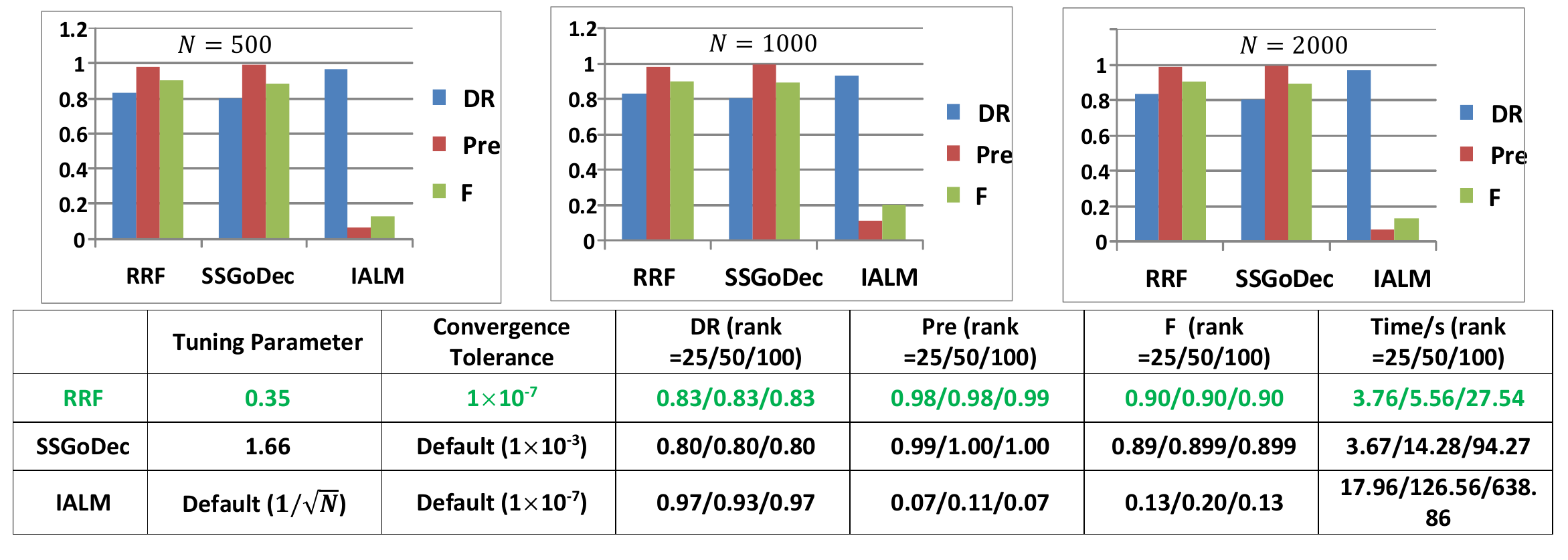}
\end{center}
\vspace{-1em}
   \caption{\label{fig:3}\textbf{Simulation Results}: Top: overall performance measured by detection rate (DR), precision (Pre) and $F$-measure under different matrix dimensions $N$. Both the proposed RRF and SSGoDec have the highest scores and IALM performs poor both in precision and F-measure. Bottom: presents the specific tuning parameter, convergence tolerance assigned to each model and its the corresponding DR, PRE, F-measure scores and the total computation time under different dimensions $N$. }
\end{figure*}

We first demonstrate the performance of the proposed model on simulated data compared to GoDec and RPCA with their fastest implementations, SSGodec and IALM (inexact ALM) respectively.
Inexact ALM \cite{lin2010augmented} has much higher computation speed and higher precisions than previous algorithm for RPCA problem.
SSGoDec \cite{SSGoDec} is also an improved version for GoDec \cite{zhou2011godec}, which largely reduces the time cost with the error non-increased.

Without loss of generality, we can construct a square matrices of three different dimensions $N = 500, 1000, 2000$ and set rank $r = 5\% N$. We also need to generate three types of matrices for simulation, that is the low rank matrix $L\in R^{N\times N}$, the dense Gaussian noise matrix $G\in R^{N\times N}$, and the Bernoulli noise matrix $B\in R^{N\times N}$.
$L$ can be generated from a random matrix $P =rand(N,r)$ by setting $ L=PP^T$, where $rand(N,r)$ will generate a $N\times r$ random matrix. The Gaussian matrix can be set as $G = 2*rand(N,N)$ with the coefficient indicating its variance level. The Bernoulli matrix can be set as $B = 10*sprand(N,N,5\%)$, where $sprand(N,N,5\%)$ will generate a sparse $N\times N$ random matrix with $5\%$ indicating its sparse level.

Experimental results will be evaluated by using the following three metrics: the detection rate (DR)/recall, the precision (Pre) and the F-measure (F) \cite{maddalena2010fuzzy}:
\begin{equation}
\begin{split}
&DR = \frac{tp}{tp+fn}\\
&Pre = \frac{tp}{tp+fp}\\
&F =\frac{2\times DR \times Pre}{DR + Pre}
\end{split}
\end{equation}
tp indicates the total number of corrupted pixels that are correctly detected; fn indicates the total number of corrupted pixels that are not being detected;
fp denotes the total number of detected pixels which are actually normal.
A good low-rank recovery or a precise sparse-errors extraction should have high detection rate, high precision and high F-measure. Among all the three metrics, F-measure is the most synthesized.

The results are presented in Fig.\ref{fig:3}. The parameters are tuned for best performances. We observed that the proposed robust rank factorization achieves very high scores of all the three metrics, which imply a very accurate recovery of low-rank matrices as well as a precise sparse-error detection. Meanwhile it pays the lowest time cost.
\subsubsection{Background Modeling}
\begin{figure}
\begin{center}
\includegraphics[width=1\linewidth]{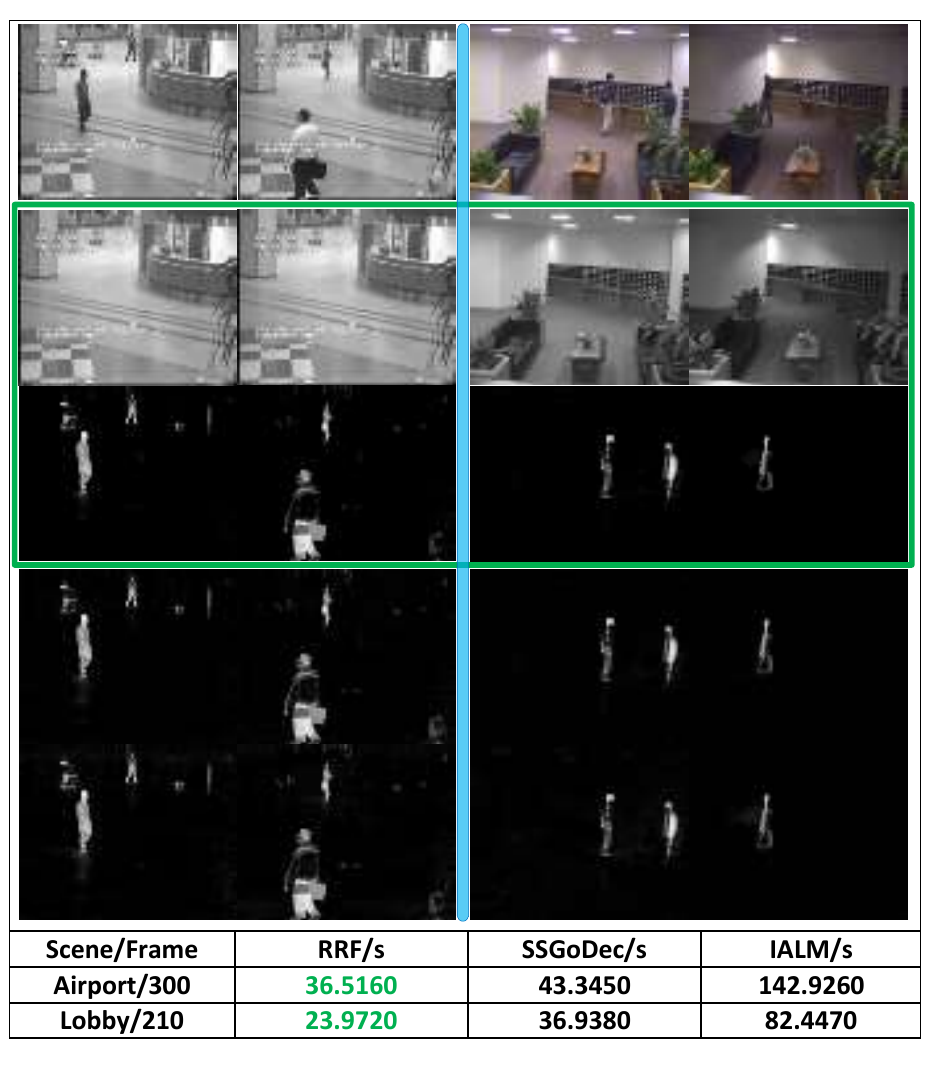}
\end{center}
\vspace{-1em}
   \caption{\label{fig:4}\textbf{Background Modeling}: Left: airport scenario. Right: lobby scenario. The first row of the large image presents two original frames of each video. The following two rows inside the green box are extracted backgrounds and foregrounds by the proposed model. The last two rows are foregrounds extracted by SSGoDec and IALM respectively. The table on the bottom gives the total running time of each model for each video.}
\end{figure}

In this part, comparison experiments on real video data are conducted to further demonstrate the high computation efficiency of the proposed rank factorization method.
We test the models on two video data from \cite{li2004statistical}: (1) airport (300 frames): there is no significant light changes in this video, but it has a lot of activities in the foreground. The image size is $144\times 176$;
(2) lobby (210 frames), there is little activity in this video, but it goes through significant illumination changes. The image size is $128\times160$.
Fig. \ref{fig:4} shows that each model achieves equally results and the proposed method remains to be the fastest among the three.

In a word, both the simulations and real-data experiments strongly validate the high computation efficiency and strong robustness of the the proposed model.

\section{Conclusion}
This paper presents a powerful regression method, LOIRE, which is efficient in detecting outliers and contributes a lot to approximately achieve a Bernoulli estimate which is more accurate than LOIRE but with an almost equivalent efficiency as LOIRE. Also, LOIRE can be further extended to realize a robust rank factorization which inherits the high efficiency of LOIRE and outperformed the state-of-the-art low-rank recovery methods, IALM and SSGoDec both on simulations and on real-data experiments in terms of efficiency with a competitive accuracy.

\appendix
\appendixpage
\addcontentsline{toc}{section}{}\markboth{APPENDIX}{}
\begin{appendix}
\noindent \textbf{Proof of Proposition 1}£ºFirst let at look at the following lemma:

\noindent \textbf{Lemma 1}: $\quad$ In problem (\ref{L0Model}), if $(x^0,e^0,b^0)$ is one of its optimal solution pair and $\mathcal{I}$ is a support set of $b^0$, i.e. $b_i \neq 0$ for $i\in \mathcal{I}$, then we should have $e^0_i= 0$ for $i\in \mathcal{I}$.

\noindent \textit{Proof of Lemma 1}: Let's consider another equivalent form of problem (\ref{L0Model}) as follows:
\begin{equation}
\begin{split}
&<x^0,e^0,b^0> = \arg\min_{x,e} \frac{\lambda}{2}\|e\|_2^2 +\|b\|_0\\
&s.t. \quad e = y-Ax-b.
\end{split}
\end{equation}
Suppose there $\exists k\in \mathcal{I}$ such that $e_k^0\neq 0$, then we could construct another feasible pair $(b^{\prime}, e^{\prime} )$ as below:
\begin{equation}
\begin{split}
&\tilde{b}_k = b_k^0 + e_k^0, \tilde{e}_k = 0\\
&\tilde{b}_{-k} = b^0_{-k}, \tilde{e}_{-k} = e^0_{-k},
\end{split}
\end{equation}
where $b_{-k}$ indicates all the entries in $b$ except $b_k$.
It is obvious that $\|\tilde{b}\|_0\le \|b^0\|$. Since $e_k^0\neq 0$, then we must have $\|e_k^0\|^2>0$ and
\begin{equation}
\begin{split}
&\|b^0\|_0 +\frac{\lambda}{2}\|e^0\|_2^2\\
&= \|b^0\|_0 +\frac{\lambda}{2}(\|e^0_{-k}\|_2^2 +\|e^0_{k}\|_2^2)\\
&\ge \|\tilde{b}\|_0  +\frac{\lambda}{2}(\|e^0_{-k}\|_2^2 +\|e^0_{k}\|_2^2)\\
&> \|\tilde{b}\|_0  +\frac{\lambda}{2}(\|\tilde{e}\|_2^2),\\
\end{split}
\end{equation}
which contradicts the optimality of feasible solution $(e^0,b^0)$. Then we prove the lemma. $\blacksquare$

Then we could continue on the proof for Proposition 1:
It is obvious that
\begin{equation}
\|b^0\|_0 = \|b^0_{\mathcal{I}}\|_0.
\end{equation}
According to Lemma 1, we should have
\begin{equation}
e^0_{\mathcal{I}} = y_{\mathcal{I}} -A_{\mathcal{I}}x^0 + b_{\mathcal{I}}^0
\end{equation}
According to the above two formula, we can have
\begin{equation}
\|b^0\|_0 +\frac{\lambda}{2}\|y-Ax^0-b^0\|_2^2 = \|b_{\mathcal{I}}^0\|_0 +\frac{\lambda}{2}\|y_{\mathcal{I}^c}-A_{\mathcal{I}^c}x^0\|_2^2
\end{equation}
Since $x^0$ is an optimal solution for problem (\ref{L0Model}), then
\begin{equation}
\begin{split}
&\|b_{\mathcal{I}}^0\|_0 +\frac{\lambda}{2}\|y_{\mathcal{I}^c}-A_{\mathcal{I}^c}x^0\|_2^2\\
&=\min_{x}\{\|b^0\|_0 +\frac{\lambda}{2}\|y-Ax-b^0\|_2^2 \}\\
&=\min_{x}\{\|b^0\|_0 + \frac{\lambda}{2}\|y_{\mathcal{I}^c}-A_{\mathcal{I}^c}x\|_2^2 +\frac{\lambda}{2}\|y_{\mathcal{I}}-A_{\mathcal{I}}x-b_{\mathcal{I}}^0\|_2^2 \}\\
&\ge \|b^0\|_0 + \frac{\lambda}{2}\min_{x}
\|y_{\mathcal{I}^c}-A_{\mathcal{I}^c}x\|_2^2\\
& \|\hat{b}\|_0 + \frac{\lambda}{2}\|y-Ax-\hat{b}\|_2^2\\
&where \quad \hat{b}_{\mathcal{I}^c} =0, \hat{b}_{\mathcal{I}} = y_{\mathcal{I}}-A_{\mathcal{I}}x^{\prime}\\
&\ge \min_{x,b}\|b\|_0 + \frac{\lambda}{2}\|y-Ax-b\|_2^2
\end{split}
\end{equation}
In the above inference, since the first row equals to the last row, therefore, all the $\ge$ should be replaced by $=$ and hence we should have $Ax^0 = Ax^{\prime}$. $\blacksquare$
\end{appendix}

{\small
\bibliographystyle{ieee}
\bibliography{egbib}
}

\end{document}